\documentclass[10pt,journal,letterpaper]{IEEEtran}
\usepackage[T1]{fontenc}
\usepackage[utf8]{inputenc}
\usepackage{cite}
\usepackage{amsmath,amssymb}
\usepackage{graphicx}
\usepackage{array,booktabs,multirow,tabularx}
\usepackage[hidelinks]{hyperref}
\hypersetup{pdftitle={iFeST: An Iterative Motion Correction Framework for Fetal Brain Diffusion MRI via Frequency-Spatial-Temporal Modeling},pdfauthor={Haoan Xu, Ruike Chen, Jing Xia, Haotian Li, Huanyun Zhang, Cong Sun, Guangbin Wang, Shuhao Zhang, Zhigang Gao, Dan Wu}}
\begin{document}
\title{iFeST: An Iterative Motion Correction Framework for Fetal Brain Diffusion MRI via Frequency-Spatial-Temporal Modeling}
\author{Haoan~Xu\textsuperscript{1,2}, Ruike~Chen\textsuperscript{1}, Jing~Xia\textsuperscript{1}, Haotian~Li\textsuperscript{1}, Huanyun~Zhang\textsuperscript{1}, Cong~Sun\textsuperscript{3}, Guangbin~Wang\textsuperscript{4}, Shuhao~Zhang\textsuperscript{5}, Zhigang~Gao\textsuperscript{5}, and Dan~Wu\textsuperscript{1,6}%
\thanks{\textsuperscript{1}Key Laboratory for Biomedical Engineering of Ministry of Education, Department of Biomedical Engineering, College of Biomedical Engineering \& Instrument Science, Zhejiang University, Hangzhou, China}%
\thanks{\textsuperscript{2}Institute of Research and Clinical Innovations, Neusoft Medical Systems Co., Ltd., Hangzhou, China}%
\thanks{\textsuperscript{3}Department of Radiology, Beijing Hospital, Beijing, China}%
\thanks{\textsuperscript{4}Department of Radiology, Shandong Provincial Hospital Affiliated to Shandong First Medical University, Jinan, China}%
\thanks{\textsuperscript{5}Department of General Surgery, Children's Hospital, Zhejiang University School of Medicine, Hangzhou, China}%
\thanks{\textsuperscript{6}Department of Electrical and Electronic Engineering, University of Nottingham Ningbo China, Ningbo, China}%
}
\maketitle
\pagestyle{plain}
\thispagestyle{plain}

\begin{abstract}
Fetal brain diffusion magnetic resonance imaging (dMRI) is a pivotal non-invasive tool for assessing the development of brain connectivity and microstructure. However, the prolonged acquisition times and complex sampling requirements render dMRI highly susceptible to unpredictable fetal motion. Existing motion correction methods typically process each dMRI volume independently and focus on inter-slice motion. Such strategies fail to utilize the coherent information between dMRI volumes along different diffusion directions and lack the ability to restore the intra-slice corruption. To address these limitations, we proposed iFeST, an iterative framework that combines frequency (k-space), spatial, and temporal (inter-slice) modeling for fetal brain dMRI motion correction. At each iteration, iFeST integrates information across diffusion directions into fiber orientation distributions (FODs). These distributions encode local fiber orientations within each voxel and serve as microstructural priors for intra-slice restoration and inter-slice alignment in the next iteration. To train the framework, we developed a fetal brain dMRI motion simulator that synthesizes compound motion artifacts from FODs. Evaluation on simulated datasets demonstrated that iFeST outperformed four inter-slice networks in motion estimation and three intra-slice networks in image restoration. Further validation on a real-world dataset confirmed significantly better volume reconstruction and FOD estimation than conventional pipelines. Overall, iFeST provides a learning-based framework for fetal dMRI motion correction, restoring motion-corrupted data and improving microstructural reconstruction for downstream diagnostic applications.
\end{abstract}

\begin{IEEEkeywords}
Fetal Brain MRI, Diffusion MRI, Motion Correction, Deep Learning
\end{IEEEkeywords}

\section{Introduction}\label{sec:1}

Fetal brain diffusion magnetic resonance imaging (dMRI) non-invasively models water diffusion properties \cite{ref1}, serving as an essential tool for quantitatively probing the development of brain connectivity and microstructure \cite{ref2,ref3,ref4}. Fundamentally, dMRI utilizes diffusion-sensitizing gradients along specific directions to encode water displacement \cite{ref5}, thereby characterizing the microstructural organization of neural tissues. With dense, redundant sampling across non-collinear gradient directions, these diffusion signals can be modeled using diffusion tensor imaging (DTI) \cite{ref6} to estimate the primary orientation of white matter tracts \cite{ref7}. High angular resolution diffusion imaging (HARDI) \cite{ref8} further supports representations of more complex fiber configurations, such as the fiber orientation distribution (FOD) \cite{ref9}. Commonly represented using spherical harmonic (SH) coefficients, the FOD captures voxel-wise fiber architecture, supporting crossing-fiber tractography and microstructural analysis.

Because dMRI is sensitive to microscopic water displacement, bulk fetal motion can also be encoded in the signal and degrade image quality. Moreover, dMRI acquisition typically uses diffusion-weighted echo-planar imaging (DW-EPI) with single- or multi-shell sampling across \(>30\) directions per shell \cite{ref10}, and thus requires a relatively long scan time. Specifically, a substantial portion of each slice acquisition (approximately 100 ms) is devoted to k-space filling. Motion during this readout disrupts phase encoding, causing intra-slice aliasing and blurring, whereas motion between adjacent slice acquisitions causes inter-slice misalignment. The resulting compound artifacts severely compromise downstream microstructural modeling and tractography, making it necessary to develop dedicated motion correction (MoCo) methods for fetal dMRI.

\subsection{Related Works}\label{sec:1.1}

\subsubsection{MoCo in Fetal dMRI}\label{sec:1.1.1}

Conventional retrospective fetal dMRI MoCo primarily relied on iterative optimization. For single-shell dMRI, the most widely adopted MoCo framework was proposed by Deprez et al. \cite{ref11} and integrated into SVRTK (\url{https://github.com/SVRTK/SVRTK}). This pipeline followed a two-stage strategy: inter-slice motion correction and SH reconstruction. The first stage was adapted from MoCo in fetal structural MRI (sMRI) \cite{ref12,ref13}, treating diffusion-weighted (DW) stacks as independent volumes and utilizing iterative slice-to-volume registration (SVR) \cite{ref14} and super-resolution reconstruction (SRR) to estimate each slice's rigid motion parameters. In the subsequent stage, all aligned slices were mapped into SH space. Forward projection adjusted slice weights to de-weight or reject intra-slice motion-corrupted slices until convergence. However, such optimization-based algorithms were computationally expensive and struggled with large-scale motion. Furthermore, by discarding corrupted data through outlier rejection, SVRTK was not designed to correct intra-slice artifacts and did not exploit dMRI data redundancy, compromising the overall reconstruction.

Subsequent research extended MoCo to multi-shell dMRI. Christiaens et al. \cite{ref15} introduced SHARD, combining angular SH and radial decomposition to reconstruct multi-shell data from scattered, motion-corrupted slices in a unified spatial-angular-radial representation. More recently, Snoussi et al. \cite{ref16} developed HAITCH, a framework designed to jointly address motion and dynamic geometric distortions. Using a model-free SHORE basis for the diffusion signal, HAITCH iterated slice weighting, basis updates, and rigid registration. Despite their advancements, these three pipelines shared fundamental limitations: they treated intra-slice artifacts primarily through outlier rejection and relied on traditional SVR optimization for inter-slice estimation. In a different direction, Verdera et al. \cite{ref17} presented HERON, a prospective MoCo scheme using deep learning for real-time fetal brain localization and motion assessment on the scanner, enabling automatic re-acquisition of motion-corrupted data.

\subsubsection{Deep Learning-Based MoCo}\label{sec:1.1.2}

Although deep learning-based retrospective MoCo for fetal brain dMRI has not been reported, MoCo in fetal sMRI was extensively studied as an inter-slice parameter estimation task \cite{ref18}. For instance, Hou et al. \cite{ref19} proposed SVR-Net, using convolutional neural networks (CNNs) with fully connected layers to regress slice-wise rigid motion parameters. Building upon CNNs, Singh et al. \cite{ref20} integrated long short-term memory (LSTM) to extract temporal information from the slice sequence. To provide explicit spatial guidance, Shi et al. \cite{ref21} proposed AFFIRM, using a 3D reference volume as a spatial prior refined through iterative reconstruction. More recently, Xu et al. \cite{ref22,ref23} presented SVoRT, leveraging Transformers to enhance temporal modeling. Their iterative and incremental predictions enabled more stable convergence toward the ground truth (GT).

Regarding intra-slice MoCo in fetal sMRI, research has been relatively limited, as the readouts are relatively fast and robust and less affected by motion. Among the few attempts, Lim et al. \cite{ref24} employed a generative adversarial network (GAN), utilizing an autoencoder generator to synthesize motion-corrected images and a CNN discriminator to assess anatomical realism. Conversely, learning-based intra-slice MoCo was more established in adult sMRI \cite{ref25}. Numerous studies utilized U-Net architectures \cite{ref26} and their variants to translate motion-corrupted images into corrected representations \cite{ref27,ref28,ref29}. To further enhance restoration fidelity, several methods incorporated prior information as auxiliary inputs \cite{ref30,ref31}. For instance, Al-masni et al. \cite{ref32} proposed PriorNet, which leveraged adjacent slices as spatial priors. Alternatively, other approaches operated directly in k-space, training networks to correct k-space errors \cite{ref33}. For a more comprehensive review of deep learning applications in general MRI motion correction, readers are referred to \cite{ref25}.

\subsection{Contributions}\label{sec:1.2}

In this work, we proposed iFeST, an \textbf{i}terative \textbf{F}r\textbf{e}quency-\textbf{S}patial-\textbf{T}emporal modeling framework for fetal brain dMRI motion correction. To exploit the inherent redundancy across diffusion directions, we designed a reference-guided iterative strategy that integrated multi-directional dMRI information through FOD fitting, providing microstructural priors for motion-corrupted stacks. These priors were progressively refined using intermediate correction outputs to guide subsequent iterations. Furthermore, spatial-frequency modeling was incorporated to suppress intra-slice artifacts from k-space inconsistencies, while temporal modeling of the slice sequence was employed for inter-slice motion estimation. To the best of our knowledge, this is the first learning-based retrospective method designed for fetal dMRI MoCo. Code is available at \url{https://github.com/zjuwulab/iFeST}. The main contributions are summarized as follows:

\noindent (1) We designed an FOD-guided iterative framework that integrates multi-directional dMRI information into subject-specific reference stacks, providing progressively refined microstructural guidance for correcting individual dMRI stacks.

\noindent (2) Within this framework, we developed a dual-stream FeST network that leverages spatial-frequency features to suppress intra-slice artifacts and spatial-temporal features to estimate inter-slice motion parameters under reference guidance.

\noindent (3) We established a comprehensive reconstruction pipeline by integrating iFeST with existing toolboxes, enabling precise fitting of high-order dMRI parameters from severely motion-corrupted datasets.

\noindent (4) We constructed a motion simulator for fetal brain dMRI for the first time, which replicates complex dMRI motion, including intra-slice aliasing, inter-slice misalignment, and signal dropout during the dMRI sensitization period.

\section{Method}\label{sec:2}

\subsection{iFeST: Overall Iterative Correction Framework}\label{sec:2.1}

In single-shell dMRI, the input comprises \(N_{\mathrm{vol}}\) anisotropic DW stacks acquired at a fixed diffusion weighting along different gradient directions. The acquisition is denoted by \(\mathbf{y}=\{y_{i,z}\}\), where \(\mathbf{y}_{i}\) denotes the stack along the \(i\)-th gradient direction, and \(y_{i,z}\) denotes the \(z\)-th slice, with \(i=1,\ldots,N_{\mathrm{vol}}\) and \(z=1,\ldots,N_{\mathrm{slice}}\).

\begin{figure*}[!t]
\centering
\includegraphics[width=0.95\textwidth]{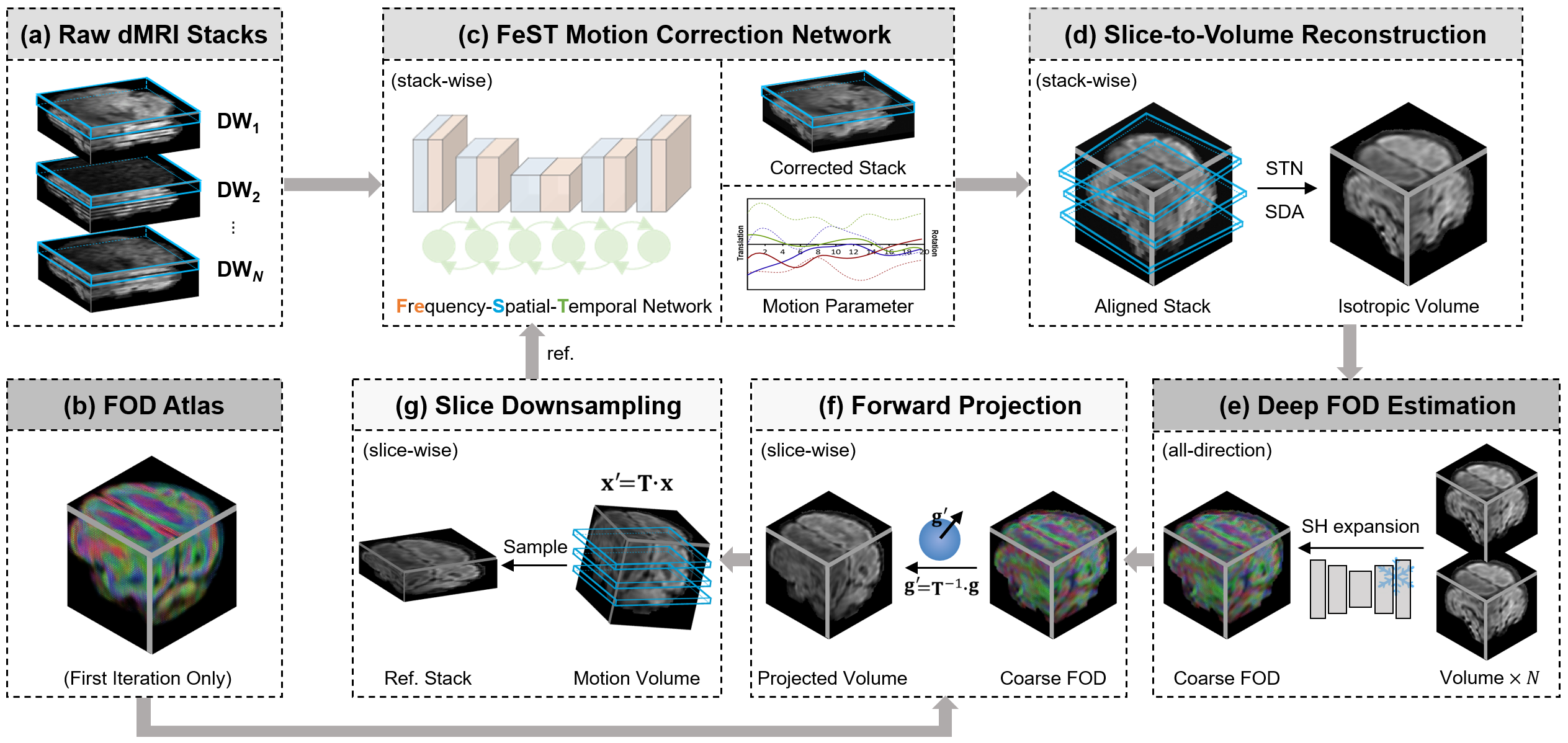}
\caption{Overview of the proposed iterative MoCo framework iFeST. The framework initializes with \(N_{\mathrm{vol}}\) dMRI stacks (a) and a GA-matched FOD atlas (b) for initial reference generation. (c) Sub-network FeST processes individual stacks, yielding intra-slice motion-corrected stacks and inter-slice motion parameters. (d) The SVR module aligns corrected slices to reconstruct an isotropic volume for each stack. (e) Volumes are aggregated to estimate FOD via a pre-trained network. Updated FOD is forward-projected in (f) and mapped to slice space in (g) to generate references for the next iteration. STN: spatial transformer network; SDA: scattered data approximation; FOD: fiber orientation distribution; SH: spherical harmonics.}
\label{fig:1}
\end{figure*}

As illustrated in Fig.~\ref{fig:1}, iFeST alternates FeST-based MoCo with FOD-based reference generation. Within each iteration, FeST processes each stack sequentially, suppressing intra-slice artifacts and estimating inter-slice motion parameters. Reference stacks are synthesized from an FOD atlas matched for gestational age (GA) \cite{ref34} in the first iteration and from the updated subject-specific FOD in subsequent iterations. Using the restored slices and estimated motion parameters, an SVR module reconstructs isotropic DW volumes, which are combined across directions for deep FOD estimation. The updated FOD is then forward-projected to synthesize more physically consistent reference stacks using the estimated slice poses and acquisition geometry. Feeding these references back into FeST establishes a coarse-to-fine loop that exploits redundancy across diffusion directions to progressively reduce artifacts.

\begin{figure*}[!t]
\centering
\includegraphics[width=0.98\textwidth]{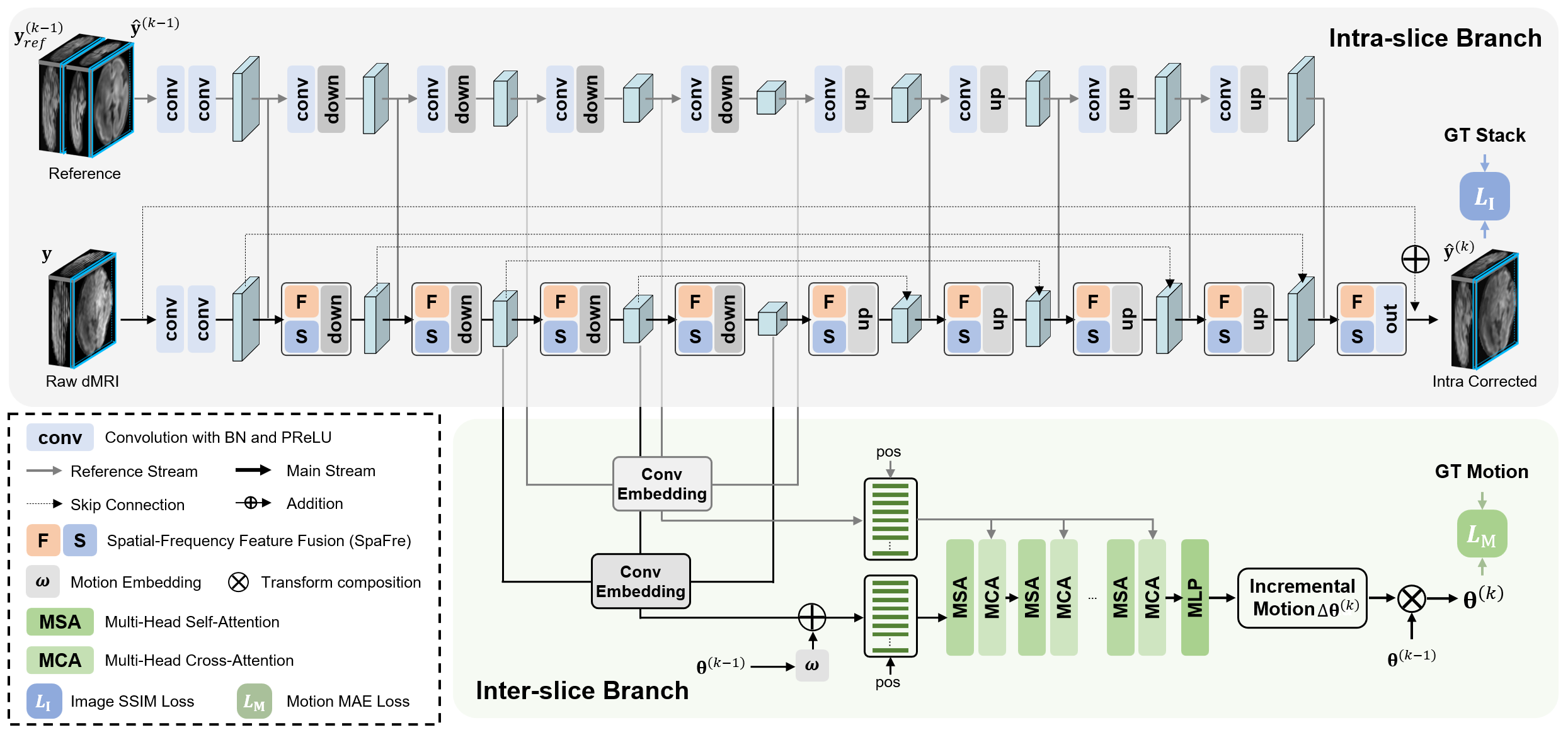}
\caption{Architecture of the FeST network. FeST comprises two parallel branches for intra-slice artifact removal and inter-slice motion estimation. The intra-slice branch takes the raw dMRI stack and a reference input comprising synthesized and previously corrected stacks. It consists of a main stream incorporating spatial-frequency feature fusion (SpaFre) modules, guided by a lightweight reference stream at each stage. Within each SpaFre module, F and S denote frequency- and spatial-domain processing, respectively. The inter-slice branch extracts hierarchical features from both streams. These features are projected into token sequences via convolutional embeddings and processed by four stacked MSA-MCA modules to capture temporal dependencies. MSA: multi-head self-attention; MCA: multi-head cross-attention.}
\label{fig:2}
\end{figure*}

\subsection{FeST Motion Correction Network}\label{sec:2.2}

\subsubsection{Network Overview}\label{sec:2.2.1}

\begin{figure*}[!t]
\centering
\includegraphics[width=0.90\textwidth]{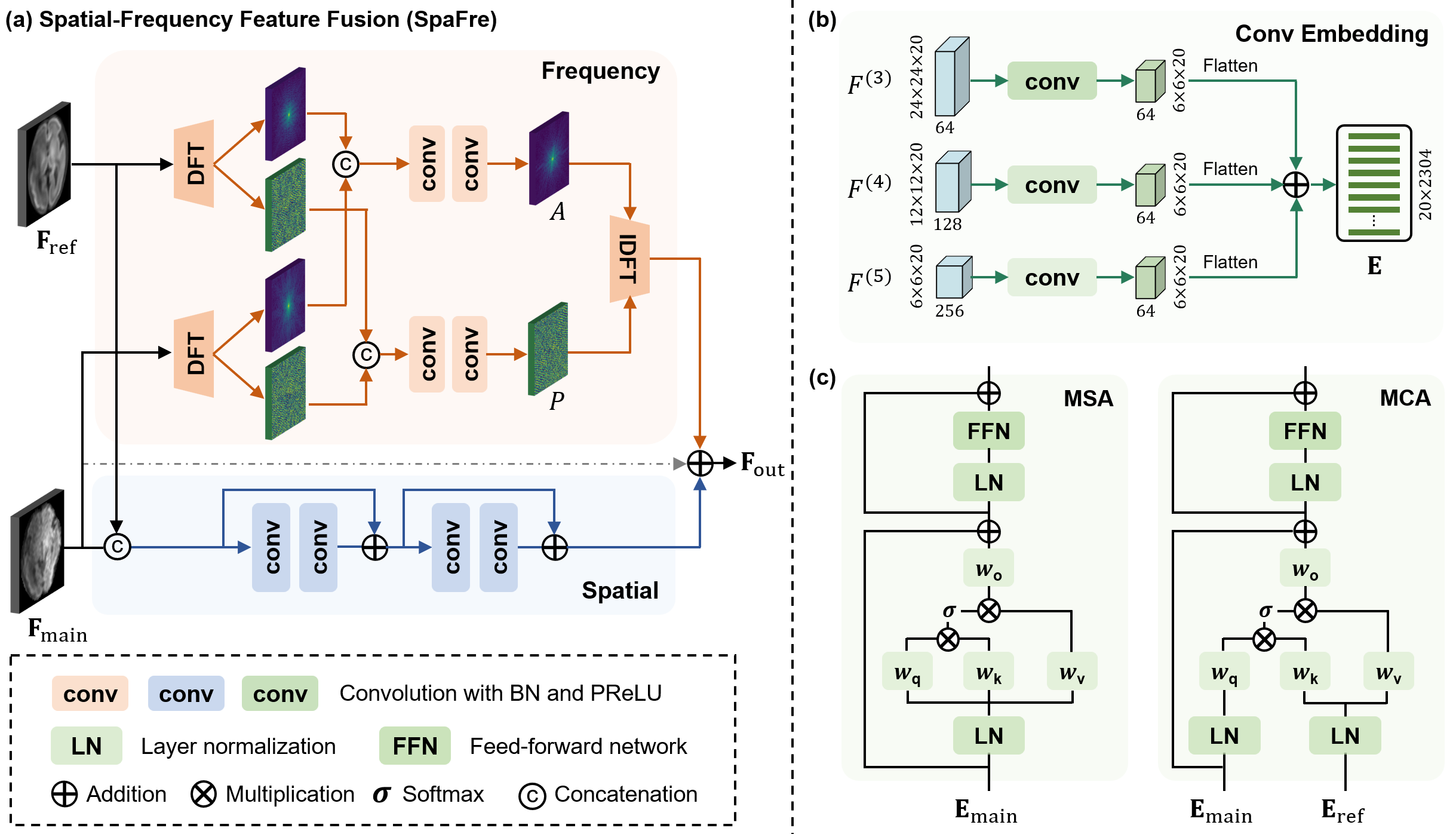}
\caption{(a) The spatial-frequency feature fusion (SpaFre) module processes reference and main features in spatial and frequency domains. The spatial path processes concatenated features using residual convolutional blocks. The frequency path applies slice-wise DFT to input features, fuses their amplitude and phase separately, and maps fused features back to the spatial domain using IDFT. (b) Convolutional embedding in the inter-slice branch. Multi-scale feature maps are projected to a unified size (\(6\times 6\)) using convolutions with kernel sizes and strides of \(4\times 4\times 1\), \(2\times 2\times 1\), and \(1\times 1\times 1\), respectively. Projected features are flattened slice-wise and summed across scales into 20 tokens. (c) Simplified schematics of the MSA and MCA blocks, with individual attention heads omitted. DFT: discrete Fourier transform; IDFT: inverse discrete Fourier transform; MSA: multi-head self-attention; MCA: multi-head cross-attention.}
\label{fig:3}
\end{figure*}

The FeST sub-network jointly suppresses intra-slice artifacts and estimates inter-slice motion increments for each DW stack. The network processes stacks sequentially to accommodate varying numbers of gradient directions across clinical protocols. Let \(\mathcal{F}\) denote the mapping function of FeST. At iteration \(k\), FeST receives the raw stack \(\mathbf{y}_{i}\), the previous motion parameters \(\boldsymbol{\theta}_{i}^{(k-1)}\), and a reference input formed by channel-wise concatenating the synthesized stack \(\mathbf{y}_{i,\mathrm{ref}}^{(k-1)}\) with the previously corrected stack \(\hat{\mathbf{y}}_{i}^{(k-1)}\), yielding the corrected stack \(\hat{\mathbf{y}}_{i}^{(k)}\) and the motion parameter increments \(\Delta\boldsymbol{\theta}_{i}^{(k)}\). The corrected stack suppresses intra-slice artifacts while retaining slice-wise poses, and inter-slice alignment is performed during subsequent reconstruction. The mapping is
\begin{equation}
\label{eq:1}
\bigl(\hat{\mathbf{y}}_{i}^{(k)},\Delta\boldsymbol{\theta}_{i}^{(k)}\bigr)=\mathcal{F}\bigl(\mathbf{y}_{i},[\mathbf{y}_{i,\mathrm{ref}}^{(k-1)},\hat{\mathbf{y}}_{i}^{(k-1)}],\boldsymbol{\theta}_{i}^{(k-1)}\bigr).
\end{equation}

For motion estimation, FeST uses incremental prediction. Let \(\mathcal{T}(\boldsymbol{\theta})\) denote the rigid transformation for the motion parameters \(\boldsymbol{\theta}\). Since the reference \(\mathbf{y}_{i,\mathrm{ref}}^{(k-1)}\) reflects the anatomical state aligned according to \(\boldsymbol{\theta}_{i}^{(k-1)}\), the predicted \(\Delta\boldsymbol{\theta}_{i}^{(k)}\) represents the residual motion relative to this reference. Consequently, the global motion parameters are updated for the current iteration by matrix composition
\begin{equation}
\label{eq:2}
\mathcal{T}\bigl(\boldsymbol{\theta}_{i}^{(k)}\bigr)=\mathcal{T}\bigl(\Delta\boldsymbol{\theta}_{i}^{(k)}\bigr)\circ\mathcal{T}\bigl(\boldsymbol{\theta}_{i}^{(k-1)}\bigr).
\end{equation}

As illustrated in Fig.~\ref{fig:2}, FeST comprises an intra-slice image-restoration branch and an inter-slice motion-regression branch. The intra-slice branch adopts an asymmetric dual-stream architecture. The main stream processes the raw stack using spatial-frequency feature fusion (SpaFre) modules \cite{ref35}, whereas the reference stream uses lightweight convolutions to extract guidance from the synthesized and previously corrected stacks. Features from the two streams are fused hierarchically. The inter-slice branch uses multi-scale features from both streams and attention mechanisms \cite{ref36} to model inter-slice dependencies and estimate motion increments.

\subsubsection{Intra-Slice Branch with Spatial-Frequency Fusion}\label{sec:2.2.2}

The intra-slice branch employs a U-shaped architecture comprising a four-stage encoder, a bottleneck, and a four-stage decoder, with a \(96\times 96\times 20\) input and 16 base channels. To accommodate anisotropic acquisition and avoid mixing across slices, feature-extraction convolutions use asymmetric \(3\times 3\times 1\) kernels, followed by batch normalization (BN) \cite{ref37} and parametric rectified linear unit (PReLU) activation \cite{ref38}. During encoding, feature maps are downsampled using convolutions with a stride of \(2\times 2\times 1\). The decoder performs upsampling through trilinear interpolation and incorporates skip connections from the encoder.

Since motion during readout introduces inconsistencies between k-space (frequency domain) lines, the SpaFre module complements spatial structural information with Fourier-domain feature modeling under reference guidance (Fig.~\ref{fig:3}(a)). Let \(\mathbf{F}_{\mathrm{main}}\) and \(\mathbf{F}_{\mathrm{ref}}\) denote the main-stream and reference-stream feature maps, respectively. The spatial path concatenates these features along the channel dimension and processes them using residual convolutional blocks to obtain \(\mathbf{F}_{\mathrm{spa}}\). In parallel, the frequency path applies a slice-wise 2D discrete Fourier transform (DFT) to map both feature maps into the frequency domain. Their amplitude and phase components are concatenated separately and processed by two independent convolutional paths. The fused amplitude and phase are recombined and mapped back to the spatial domain via a 2D inverse DFT (IDFT), yielding \(\mathbf{F}_{\mathrm{fre}}\). Residual fusion yields \(\mathbf{F}_{\mathrm{out}}=\mathbf{F}_{\mathrm{main}}+\mathbf{F}_{\mathrm{spa}}+\mathbf{F}_{\mathrm{fre}}\).

\subsubsection{Inter-Slice Branch}\label{sec:2.2.3}

The inter-slice branch receives multi-scale features from both streams at the third and fourth encoder stages and the bottleneck, corresponding to feature levels 3, 4, and 5 (Figs.~\ref{fig:2} and~\ref{fig:3}(b)). The feature maps have in-plane sizes of \(24\times 24\), \(12\times 12\), and \(6\times 6\) and channel numbers of 64, 128, and 256, respectively. Convolutions with kernel sizes and strides of \(4\times 4\times 1\), \(2\times 2\times 1\), and \(1\times 1\times 1\) project these maps to a common in-plane size of \(6\times 6\) with 64 channels. The projected features are flattened slice-wise and summed across scales, yielding one 2304-dimensional token per slice:
\begin{equation}
\label{eq:3}
\mathbf{E}=\sum_{l=3,4,5}\operatorname{Flatten}\bigl(\operatorname{conv}^{(l)}(\mathbf{F}^{(l)})\bigr).
\end{equation}

This embedding operation is applied separately to the main and reference streams to obtain \(\mathbf{E}_{\mathrm{main}}\) and \(\mathbf{E}_{\mathrm{ref}}\). A learnable positional embedding \(\mathbf{E}_{\mathrm{pos}}\) is added to both streams to encode slice order. To support incremental prediction, the motion parameters from the previous iteration are linearly projected and added to the main-stream embedding through a learnable gating coefficient \(\alpha\):
\begin{equation}
\label{eq:4}
\mathbf{E}_{\mathrm{main}}'=\mathbf{E}_{\mathrm{main}}+\mathbf{E}_{\mathrm{pos}}+\alpha\cdot\boldsymbol{w}_{\mathrm{proj}}\bigl(\boldsymbol{\theta}^{(k-1)}\bigr),
\end{equation}
\begin{equation}
\label{eq:5}
\mathbf{E}_{\mathrm{ref}}'=\mathbf{E}_{\mathrm{ref}}+\mathbf{E}_{\mathrm{pos}}.
\end{equation}

The resulting embeddings are processed by four stacked MSA-MCA modules (Figs.~\ref{fig:2} and~\ref{fig:3}(c)). Within each module, MSA models inter-slice dependencies in the main-stream embedding. MCA then uses the updated main-stream embedding as queries and the reference-stream embedding as keys and values to estimate motion residuals based on positional priors. Attention uses eight heads, and the feed-forward network (FFN) has a hidden dimension of 4096 and a dropout rate of 0.2. Finally, independent linear regressors produce 6-dimensional motion increments \(\boldsymbol{\Delta\theta}^{(k)}\) for each slice.

\subsection{FOD-Based Reference Generation}\label{sec:2.3}

Following FeST, the corrected DW stacks and estimated motion parameters are used to construct subject-specific references for the subsequent iteration (Fig.~\ref{fig:1}(d)--(g)).

\subsubsection{Slice-to-Volume Reconstruction}\label{sec:2.3.1}

For reconstruction, a simplified slice acquisition model relates the observed slices to the latent high-resolution isotropic volume, assuming a fixed rigid pose during each slice acquisition. The imaging process for the \(z\)-th observed slice \(\mathrm{y}_{i,z}\) is modeled as the latent volume \(\mathbf{x}_{i}\) subjected to rigid motion \(\boldsymbol{\theta}_{i,z}\), point spread function (PSF) blurring \(P\), and slice sampling \(\mathcal{D}_{z}\):
\begin{equation}
\label{eq:6}
y_{i,z}=\mathcal{D}_{z}P\mathcal{T}(\boldsymbol{\theta}_{i,z})\mathbf{x}_{i}+\boldsymbol{\varepsilon}_{i,z},
\end{equation}
where \(\mathcal{T}(\boldsymbol{\theta}_{i,z})\) represents the transformation matrix corresponding to \(\boldsymbol{\theta}_{i,z}\), and \(\boldsymbol{\varepsilon}_{i,z}\) denotes acquisition noise. This model does not explicitly represent intra-slice motion. At each iteration \(k\), a GPU-accelerated SVR module based on scattered data approximation (SDA) reconstructs an isotropic DW volume \(\mathbf{x}_{i}^{(k)}\) from the intra-slice-corrected \(\hat{\mathbf{y}}_{i}^{(k)}\) and estimated motion parameters \(\boldsymbol{\theta}_{i}^{(k)}\) provided by FeST (Fig.~\ref{fig:1}(d)) \cite{ref23}. Specifically, the corrected slices and their masks are convolved with a Gaussian PSF and mapped into the standard coordinate space according to the estimated motion. Contributions from overlapping slices are accumulated and normalized by the corresponding sampling support.

\subsubsection{Deep FOD Estimation}\label{sec:2.3.2}

The reconstructed DW volumes are combined across diffusion directions to estimate the FOD (Fig.~\ref{fig:1}(e)). We first express the voxel-wise normalized diffusion signal using a compact SH representation \cite{ref39}. For a gradient direction \(\mathbf{g}(\vartheta,\phi)\), the signal is approximated as:
\begin{equation}
\label{eq:7}
S(\vartheta,\phi)\approx\sum_{l=0}^{L_{\mathrm{max}}}\sum_{m=-l}^{l}c_{l,m}^{(S)}Y_l^m(\vartheta,\phi),
\end{equation}
where \(Y_l^m\) denotes the SH basis function of degree \(l\) and order \(m\) \cite{ref40}, \(c_{l,m}^{(S)}\) denotes the diffusion-signal coefficient, and \(L_{\mathrm{max}}\) specifies the truncation level.

The diffusion signal is related to the FOD through spherical convolution with a single-fiber response function. In the SH domain, this relationship is expressed as
\begin{equation}
\label{eq:8}
c_{l,m}^{(S)}=c_{l,m}^{(\mathrm{FOD})}\cdot R_l,
\end{equation}
where \(c_{l,m}^{(\mathrm{FOD})}\) denotes the FOD coefficient and \(R_l\) denotes the degree-wise response factor. Conventional constrained spherical deconvolution (CSD) estimates FOD coefficients through constrained inversion of this relationship \cite{ref41}. In this study, CSD is used for FOD reconstruction during data preparation and downstream analysis with MRtrix3 \cite{ref42}. Within the iFeST loop, a pre-trained lightweight 3D U-Net instead provides a faster feed-forward mapping from diffusion-signal SH coefficients to FOD coefficients \cite{ref43}. Its parameters remain frozen during iFeST training and inference.

\subsubsection{Reference Slice Sampling}\label{sec:2.3.3}

Given the estimated FOD, forward projection synthesizes the normalized DW signal at each voxel along a target gradient \(\mathbf{g}\) by evaluating its response-weighted SH expansion:
\begin{equation}
\label{eq:9}
x=\sum_{l=0}^{L_{\mathrm{max}}}\sum_{m=-l}^{l}\bigl(c_{l,m}^{(\mathrm{FOD})}\cdot R_l\bigr)\cdot Y_l^m(\mathbf{g}).
\end{equation}

Because slice-wise fetal rotation changes the orientation of the diffusion-encoding gradient relative to the underlying brain tissue, the gradient direction for the \(z\)-th slice of the \(i\)-th stack is adjusted using the inverse of the estimated rotation:
\begin{equation}
\label{eq:10}
\mathbf{g}_{i,z}'=\mathbf{R}(\boldsymbol{\theta}_{i,z})^{-1}\cdot\mathbf{g}_{i},
\end{equation}
where \(\mathbf{R}(\boldsymbol{\theta}_{i,z})\) denotes the rotational component of the estimated rigid transformation. Using the forward-projection model defined in Eq.~\eqref{eq:9}, the estimated FOD is then projected along \(\mathbf{g}_{i,z}'\) to synthesize the instantaneous DW volume \(\mathbf{x}_{i,z}\).

Following forward projection, the instantaneous DW volume is mapped to the acquisition space using \(\mathbf{x}_{i,z}'=\mathcal{T}(\boldsymbol{\theta}_{i,z})\cdot\mathbf{x}_{i,z}\). The transformed volume is then blurred by the PSF and sampled at the original slice locations to generate the reference slice. Repeating this for all slices produces reference stacks consistent with the estimated acquisition geometry and FOD. These stacks are fed back into FeST to guide the next correction iteration.

\begin{figure*}[!t]
\centering
\includegraphics[width=1\textwidth]{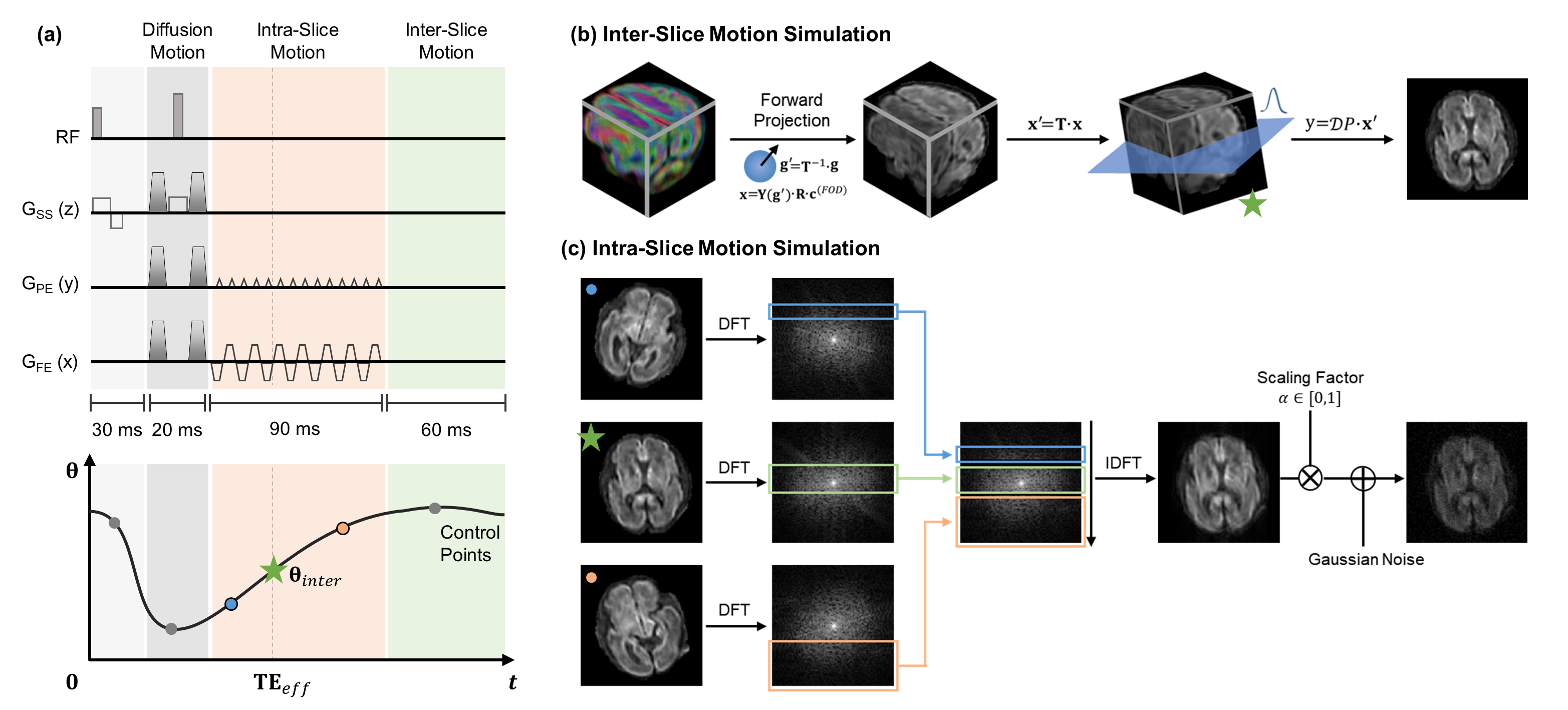}
\caption{Illustration of the motion simulation pipeline. (a) The standard EPI sequence for a single 2D slice and simulated motion trajectory. Three motion intervals are shown within the 200-ms window. The trajectory is generated via interpolation of control points (gray dots). The motion state at the effective echo time (denoted by the green star) defines the rigid transformation for inter-slice simulation, while instantaneous states during the readout window (denoted by circles) drive the intra-slice simulation. (b) Inter-slice simulation for a single slice. (c) Intra-slice simulation. The k-space of this slice is synthesized by assembling segments from different motion states. DFT: discrete Fourier transform; IDFT: inverse discrete Fourier transform.}
\label{fig:4}
\end{figure*}

\subsection{Loss Function}\label{sec:2.4}

FeST is trained with joint motion and image supervision. At each iteration \(k\), the objective adds the mean absolute error (MAE) between estimated and GT motion parameters to one minus the structural similarity index (SSIM) between corrected and GT stacks. The paired training stacks and motion targets are generated using the simulator described in Section~\ref{sec:3.2}. To prioritize the final iteration, the total loss \(\mathcal{L}\) is computed as a weighted sum across all iterations, with progressively increasing weights \((\omega_1,\omega_2,\omega_3)=(1,2,5)\):
\begin{equation}
\label{eq:11}
\begin{aligned}
\mathcal{L}&=\sum_{k=1}^{3}\omega_k\Bigl[\left\lVert\boldsymbol{\theta}^{(k)}-\boldsymbol{\theta}_{\mathrm{GT}}\right\rVert_1\\
&\qquad {}+\bigl(1-\operatorname{SSIM}(\hat{\mathbf{y}}^{(k)},\mathbf{y}_{\mathrm{GT}})\bigr)\Bigr].
\end{aligned}
\end{equation}

\section{Experiments}\label{sec:3}

\subsection{Data acquisition and preprocessing}\label{sec:3.1}

With Institutional Review Board approval, 182 fetal brain dMRIs (GA: 21.4--40.0 weeks) were acquired in axial view. Written informed consent was obtained from all pregnant volunteers. Imaging used a DW-EPI sequence on a 3T scanner (MAGNETOM Skyra, Siemens Healthcare), with b-value = 600 s/mm\textsuperscript{2} along at least 30 diffusion gradient directions and 8 non-DW images (b0). Acquisition parameters were in-plane resolution = \(1.73\times 1.73\) mm\textsuperscript{2}, slice thickness = 4 mm, repetition time (TR) = 3900 ms, echo time (TE) = 80 ms, partial Fourier factor = 6/8, and GRAPPA factor = 2.

To reconstruct high-quality FODs, raw dMRI data were preprocessed using SVRTK \cite{ref11} and MRtrix3 \cite{ref42} (\url{https://www.mrtrix.org/}). First, stacks with severe signal loss or extreme motion were excluded. Brain masks were manually delineated on b0 stacks and dilated by several voxels to encompass DW stacks. MRtrix3 preprocessing included Gibbs-ringing artifact removal \cite{ref44}, volume-to-volume alignment, eddy-current distortion correction \cite{ref45}, and bias field correction \cite{ref46}. Subsequently, SVRTK performed MoCo and SH-based SRR. Finally, FODs were computed using CSD in MRtrix3. After excluding 82 cases with failed reconstruction or poor image quality, we retained 100 subjects (GA 21.5--38.0 weeks).

In the proposed iFeST framework, a GA-matched spatiotemporal FOD atlas \cite{ref34} (covering 24--38 weeks) was utilized to provide priors for the first iteration \cite{ref47}. For subjects with a GA below the atlas range (i.e., \(<24\) weeks), the atlas at 24 weeks was adopted. The processed FODs were then randomly split into training (n=60), validation (n=15), and testing (n=25) sets. These high-quality FODs were used for subsequent motion simulation to generate paired motion-corrupted inputs and GTs.

\subsection{Motion Simulation}\label{sec:3.2}

The FOD-based simulator synthesizes DW stacks with compound motion artifacts using a DW-EPI timing model (Fig.~\ref{fig:4}). Consistent with clinical protocols, each slice has a 200-ms simulation window, an effective TE of 80 ms, and a partial Fourier factor of 6/8. Within this window, we model the effects of motion during three intervals:

\noindent (1) \textbf{Motion} \textbf{during} \textbf{diffusion} \textbf{sensitization} (30--50 ms): Motion occurring during the application of diffusion gradients induces incoherent spin dephasing, resulting in severe signal dropout.

\noindent (2) \textbf{Intra-slice motion} (50--140 ms): Motion during the EPI readout disrupts the consistency of k-space sampling, introducing intra-slice aliasing and blurring.

\noindent (3) \textbf{Inter-slice motion} (140--200 ms): Motion during the acquisition interval shifts the anatomical position of the fetus, leading to inter-slice misalignment.

\subsubsection{Motion Trajectory Generation}\label{sec:3.2.1}

We model fetal motion as a time-varying 6-degree-of-freedom (6-DoF) \cite{ref19} rigid transformation \(\boldsymbol{\theta}\), comprising 3 translation parameters \(\boldsymbol{d}=(d_x,d_y,d_z)^{T}\) and 3 rotation parameters \(\boldsymbol{r}=(r_x,r_y,r_z)^{T}\). The corresponding transformation matrix is \(\mathcal{T}(\boldsymbol{\theta})=\begin{bmatrix}\mathbf{R}(\boldsymbol{\theta})&\boldsymbol{d}\\0&1\end{bmatrix}\), where \(\mathbf{R}(\boldsymbol{\theta})\) is the rotation matrix of \(\boldsymbol{r}\). For each stack, a continuous trajectory at 1-ms resolution is generated from random-walk control points \cite{ref48}, followed by cubic spline interpolation \cite{ref49} to ensure smoothness. To match interleaved acquisition, trajectory segments are rearranged in odd-even slice order. Finally, to prevent extreme motion, rotations and translations are bounded to \(\pm\)45\(^{\circ}\) and \(\pm\)8 mm, respectively.

\subsubsection{Inter-slice Motion Simulation}\label{sec:3.2.2}

For each stack, a diffusion gradient direction is first randomly sampled. In this part, we simulate inter-slice motion-corrupted dMRI slices from FODs using the forward projection described in Section~\ref{sec:2.3.3}. For the \(z\)-th slice in the \(i\)-th stack, the simulation involves three steps. First, the motion parameters \(\boldsymbol{\theta}_{i,z}=[\boldsymbol{d}_{i,z},\boldsymbol{r}_{i,z}]\) at the effective TE are extracted from the motion trajectory. Second, the instantaneous dMRI volume is generated. While the diffusion gradient \(\mathbf{g}_{i}\) remains fixed in the scanner frame, the fetal rotation \(\boldsymbol{r}_{i,z}\) alters the relative orientation of the brain's microstructure. Consequently, the effective gradient direction \(\mathbf{g}_{i,z}'\) is computed by inversely rotating the original gradient: \(\mathbf{g}_{i,z}'=\mathbf{R}(\boldsymbol{\theta}_{i,z})^{-1}\cdot\mathbf{g}_{i}\). Using the standard-space FOD and this corrected gradient, the instantaneous isotropic dMRI volume \(\mathbf{x}_{i,z}\) is synthesized via forward projection in Eq.~\eqref{eq:9}. Third, this volume is transformed to the individual space using the rigid transformation: \(\mathbf{x}_{i,z}'=\mathcal{T}(\boldsymbol{\theta}_{i,z})\cdot\mathbf{x}_{i,z}\), and the slice is sampled from \(\mathbf{x}_{i,z}'\) based on its slicing coordinates. This procedure is repeated for all slices to simulate the stack \(\mathbf{y}_{\mathrm{inter}}\) with inter-slice motion.

\subsubsection{Intra-slice Motion Simulation}\label{sec:3.2.3}

To simulate intra-slice artifacts, a segmented k-space filling strategy is employed \cite{ref25}. The readout window is divided into several continuous sub-segments. For each sub-segment \(n\), the motion parameter \(\boldsymbol{\theta}_{n}\) at its midpoint is used to generate a slice \(\mathbf{y}_{n}\) via the inter-slice simulator. Then, FFT is applied to obtain the k-space \(\mathcal{K}_{n}\). The final intra-slice motion-corrupted k-space \(\mathcal{K}\) is synthesized by assembling the phase encoding lines from each \(\mathcal{K}_{n}\):
\begin{equation}
\label{eq:12}
\mathcal{K}(k_y,k_x)=\sum_{n=1}^{N}\mathbb{I}_{n}(k_y)\cdot\mathcal{K}_{n}(k_y,k_x),
\end{equation}
where \(\mathbb{I}_{n}\) is an indicator function that is active only for the phase encoding lines corresponding to the \(n\)-th segment. IFFT is then applied to reconstruct the slice. Intensity attenuation (with a random global scaling factor sampled uniformly from \([0,1]\)) and Gaussian noise with random variance are then applied to approximate the signal degradation associated with motion during diffusion encoding, yielding \(\mathbf{y}_{\mathrm{intra}}\). In this work, \(\mathbf{y}_{\mathrm{intra}}\) serves as the network input, while \(\mathbf{y}_{\mathrm{inter}}\) and \(\boldsymbol{\theta}\) serve as the image and motion GTs, respectively.

\subsection{Comparison Methods}\label{sec:3.3}

We first compared iFeST with state-of-the-art (SOTA) deep learning networks on simulated data. Because no existing learning-based method jointly corrects inter- and intra-slice motion, we evaluated the two tasks independently. For inter-slice motion estimation, we selected four fetal sMRI networks, SVR-Net \cite{ref19}, DeepPMT \cite{ref20}, AFFIRM \cite{ref21}, and SVoRT \cite{ref23}. For intra-slice artifact correction, we compared U-Net \cite{ref26}, PriorNet \cite{ref32}, and cGAN \cite{ref24}. To ensure a fair comparison, all competing networks were trained from scratch using the identical dataset and motion simulation. All networks were re-implemented based on their official open-source repositories, preserving original internal hyper-parameters. SVR-Net, U-Net and PriorNet were trained in a 2D configuration, whereas the remaining networks were trained in 3D. Beyond MoCo accuracy, we evaluated downstream reconstruction and microstructural analysis. Specifically, the intra-slice corrected stacks and inter-slice motion parameters were processed using SVRTK to reconstruct isotropic DW volumes, followed by diffusion modeling with MRtrix3 to derive diffusion metrics for evaluation.

For inter-slice motion estimation, we computed the weighted MAE (wMAE), Pearson correlation coefficient (\(r\)), and root mean square error (RMSE). Each slice's wMAE weight was its mask area divided by the stack's largest mask area. For intra-slice artifact correction and dMRI volume reconstruction, image fidelity was assessed using the structural similarity index (SSIM), normalized root mean square error (NRMSE), and peak signal-to-noise ratio (PSNR). Finally, the accuracy of FOD fitting was evaluated by calculating the SSIM of the zero-order SH coefficient (SH0-SSIM), the SSIM of forward-projected DW volumes (DWI-SSIM), and the angular correlation coefficient (ACC).

\subsection{Real-World Evaluation}\label{sec:3.4}

To evaluate real-world generalizability, we selected 179 DW stacks with severe inter-slice motion or signal loss from 26 subjects in the same acquisition cohort. For each subject, we selected 4 to 15 motion-corrupted directions and additional motion-free directions from the same acquisition to form a 30-direction input.

After preprocessing, stack-specific brain masks were generated using an nnU-Net-based segmentation network \cite{ref50}. The masked stacks were processed by iFeST, and its outputs were subsequently fed into the post-processing pipeline in Section~\ref{sec:3.3} for isotropic volume reconstruction and FOD estimation. The SVRTK baseline applied the same pipeline directly to preprocessed raw data. The pseudo-GT references were generated from the same subjects' redundant motion-free directions (\(\geq 30\)).

\subsection{Implementation Details}\label{sec:3.5}

\begin{table*}[!t]
\caption{Inter-slice motion estimation on simulated data. Values are reported as mean \(\pm\) standard deviation. Best values are in bold. Significant differences (\(p<0.001\)) are found between iFeST and other networks using paired t-tests.}
\label{tab:1}
\centering
\fontsize{8.5}{10.2}\selectfont
\setlength{\tabcolsep}{3pt}
\renewcommand{\arraystretch}{1.14}
\begin{tabular*}{0.98\textwidth}{@{\extracolsep{\fill}}ccccccc@{}}
\toprule
\multirow{2}{*}[-0.45ex]{Network} & \multicolumn{3}{c}{Translation} & \multicolumn{3}{c}{Rotation} \\
\cmidrule(lr){2-4}\cmidrule(lr){5-7}
 & wMAE\(\downarrow\) & \(r{\uparrow}\) & RMSE\(\downarrow\) & wMAE\(\downarrow\) & \(r{\uparrow}\) & RMSE\(\downarrow\) \\
\midrule
SVR-Net & \(1.7153\pm0.0940\) & \(0.7659\pm0.0347\) & \(2.2045\pm0.1313\) & \(3.4031\pm0.5159\) & \(0.9259\pm0.0390\) & \(4.3247\pm0.7063\) \\
DeepPMT & \(1.1838\pm0.1381\) & \(0.9026\pm0.0265\) & \(1.4357\pm0.1783\) & \(3.5290\pm0.6939\) & \(0.9036\pm0.0450\) & \(4.3990\pm0.8984\) \\
AFFIRM & \(1.0626\pm0.1418\) & \(0.9195\pm0.0219\) & \(1.3034\pm0.1794\) & \(3.5703\pm0.6666\) & \(0.9047\pm0.0428\) & \(4.4132\pm0.8174\) \\
SVoRT & \(1.0560\pm0.1655\) & \(0.9231\pm0.0224\) & \(1.2801\pm0.2107\) & \(3.1655\pm0.7890\) & \(0.9207\pm0.0488\) & \(3.9444\pm0.9973\) \\
iFeST & \(\boldsymbol{0.9100\pm0.2135}\) & \(\boldsymbol{0.9481\pm0.0283}\) & \(\boldsymbol{1.1234\pm0.3208}\) & \(\boldsymbol{2.5517\pm0.6472}\) & \(\boldsymbol{0.9501\pm0.0313}\) & \(\boldsymbol{3.1333\pm0.8862}\) \\
\bottomrule
\end{tabular*}
\end{table*}

\begin{figure*}[!t]
\centering
\includegraphics[width=0.94\textwidth]{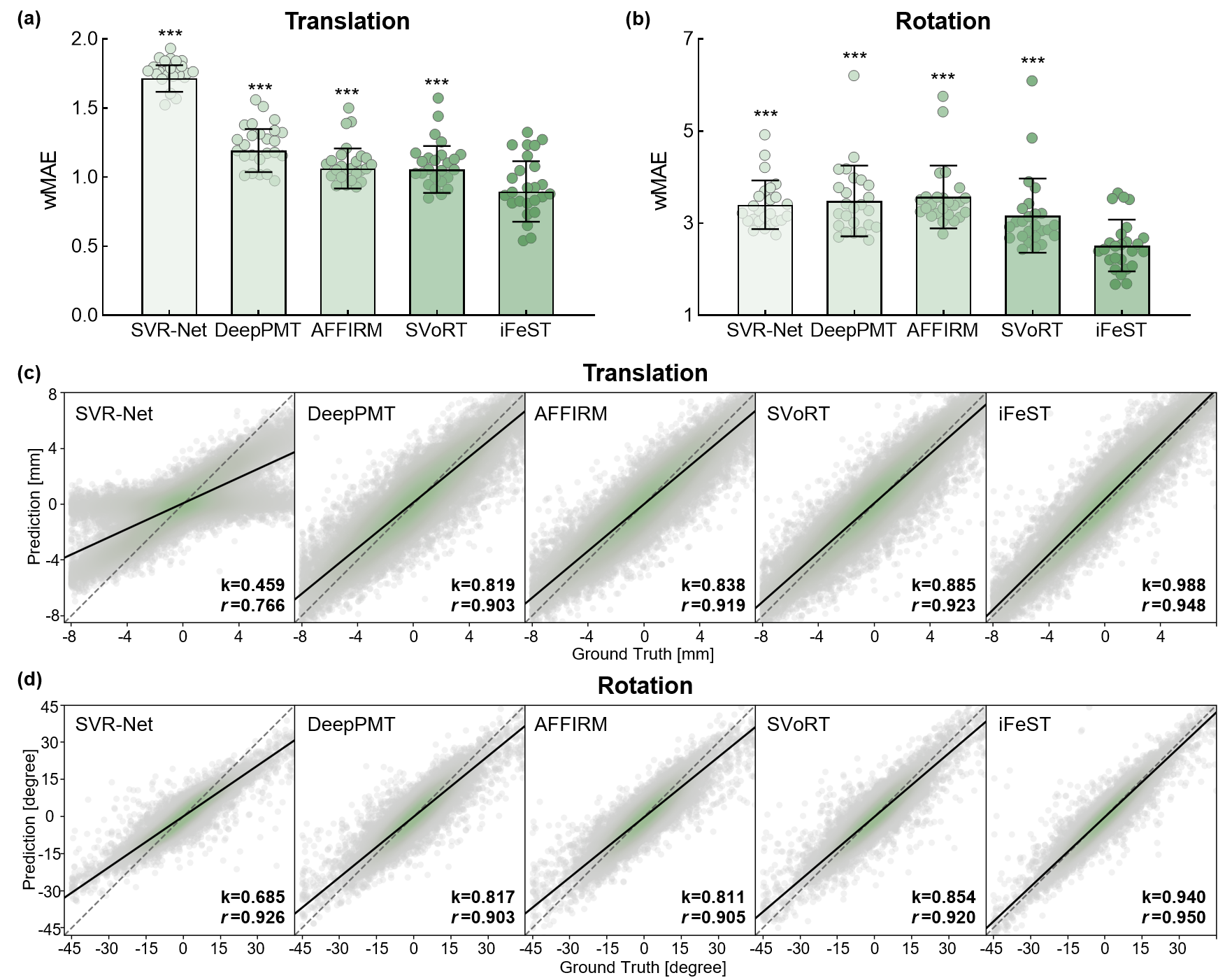}
\caption{Visual comparison of inter-slice motion estimation on the simulated dataset. (a) and (b) present the wMAE for translation and rotation, respectively. The scatter dots represent individual test cases, and asterisks indicate statistically significant differences compared to iFeST (***\(p<0.001\), paired t-test). (c) and (d) display the density scatter plots and linear regression analyses of predicted parameters versus GT. The dotted line represents the ideal agreement, while the solid line indicates the linear fit of the predictions. Here, \(k\) denotes the slope and \(r\) denotes the Pearson correlation coefficient.}
\label{fig:5}
\end{figure*}

Training pairs were generated using the motion simulator with 50-fold augmentation. To optimize training efficiency, each sample contained 6 DW stacks along random gradient directions (single-shell, b=600 s/mm\textsuperscript{2}). Validation and testing data were simulated once per subject using the full 30-direction protocol. All stacks had a \(96\times 96\) matrix, 20 slices, and 4-mm slice thickness.

The PyTorch framework was trained on an NVIDIA GeForce RTX 3090 GPU with batch size 3. iFeST was trained end-to-end using AdamW at an initial learning rate of \(3\times 10^{-5}\). A learning rate decay factor of 0.9 was applied if there was no improvement on the validation set for 5 epochs. Training lasted 200 epochs, with 60 training iterations per epoch. iFeST used 3 correction iterations and SH order 2.

\begin{figure*}[!t]
\centering
\includegraphics[width=0.96\textwidth]{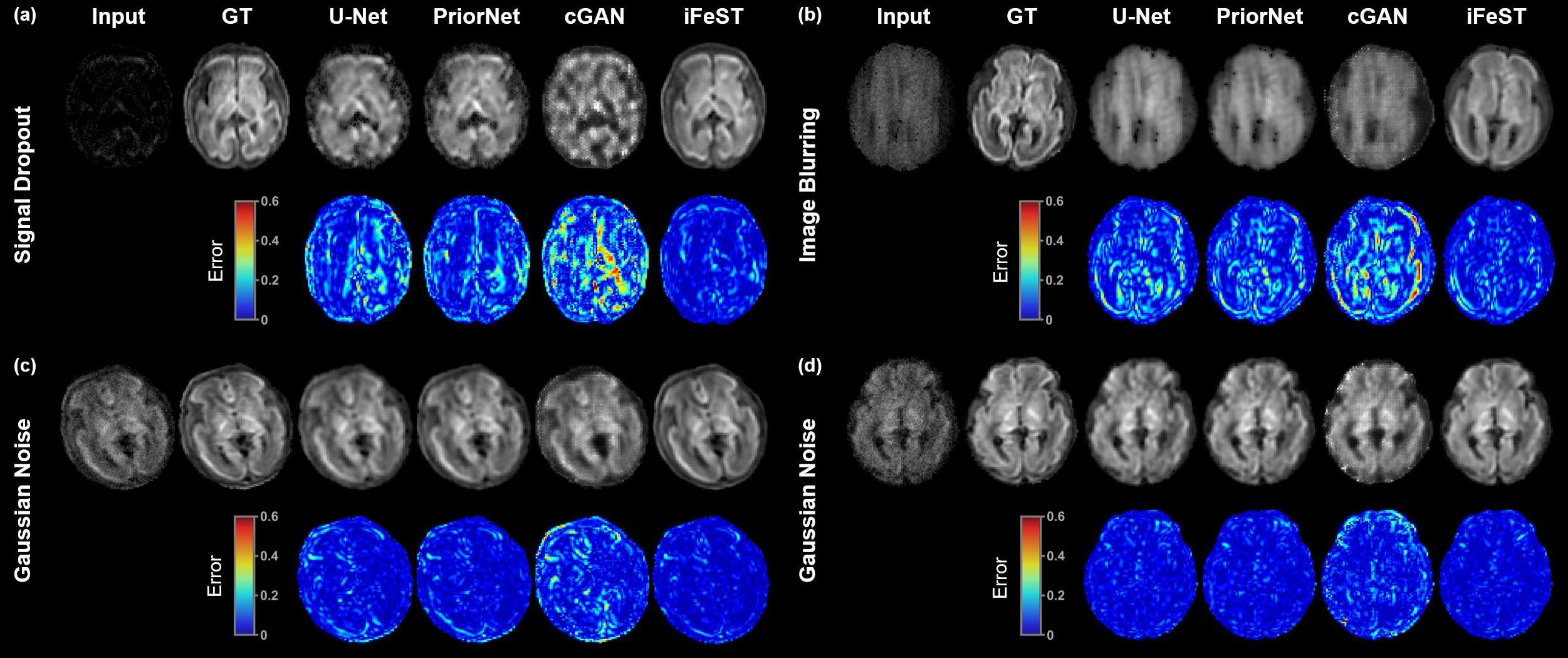}
\caption{Intra-slice motion-corrected outputs and error maps under three corruption scenarios: (a) severe signal dropout caused by diffusion encoding motion, (b) aliasing and blurring caused by intra-slice motion, and (c) and (d) low-quality images with strong noise.}
\label{fig:6}
\end{figure*}

\begin{table}[!htbp]
\caption{Intra-slice motion artifact correction on simulated data. Values are reported as mean \(\pm\) standard deviation. Best values are in bold. Significant differences (\(p<0.001\)) are found between iFeST and other networks using paired t-tests.}
\label{tab:2}
\centering
\fontsize{8.5}{10.2}\selectfont
\setlength{\tabcolsep}{3pt}
\renewcommand{\arraystretch}{1.14}
\begin{tabular}{@{}lccc@{}}
\toprule
Network & SSIM\(\uparrow\) & NRMSE\(\downarrow\) & PSNR (dB)\(\uparrow\) \\
\midrule
U-Net & \(0.8630\pm0.0366\) & \(0.0521\pm0.0063\) & \(28.33\pm0.97\) \\
PriorNet & \(0.8707\pm0.0357\) & \(0.0638\pm0.0060\) & \(26.60\pm0.56\) \\
cGAN & \(0.7870\pm0.0527\) & \(0.0715\pm0.0074\) & \(25.57\pm0.91\) \\
iFeST & \(\boldsymbol{0.8843\pm0.0368}\) & \(\boldsymbol{0.0437\pm0.0067}\) & \(\boldsymbol{29.89\pm1.17}\) \\
\bottomrule
\end{tabular}
\end{table}

\section{Results}\label{sec:4}

\subsection{Simulation Experiment}\label{sec:4.1}

\subsubsection{Inter-Slice Motion Estimation}\label{sec:4.1.1}

Table~\ref{tab:1} presents the quantitative comparison of inter-slice motion parameter estimation in the simulated dataset. Compared with the suboptimal SVoRT, iFeST reduced translation and rotation wMAE by 13.8\% and 19.4\%, respectively (\(p<0.001\), paired t-tests). It also achieved the lowest RMSE and highest \(r\). Fig.~\ref{fig:5}(a)--(b) visualize the wMAE distribution across individual test samples. To further evaluate the regression capability of each network, Fig.~\ref{fig:5}(c)--(d) plot the scatter density distributions of the predicted motion parameters versus GT for the central 10 slices of each stack (comprising 7,500 slices per sub-figure), alongside their linear fitting curves. In translation motion prediction, SVR-Net exhibited the worst performance (\(r=0.766\)) due to the absence of temporal modeling. In contrast, networks with sequence modeling demonstrated marked improvements. Among all evaluated networks, iFeST achieved the most accurate fitting (\(r=0.948\) and 0.950), demonstrating the superior parameter regression capability enabled by its spatial-frequency-temporal modeling.

\subsubsection{Intra-Slice Artifact Correction}\label{sec:4.1.2}

\begin{table*}[!t]
\caption{dMRI volume reconstruction and FOD fitting on simulated data. iFeST supplied each single-task network's complementary correction. Values are reported as mean \(\pm\) standard deviation. Best values are in bold. Significant differences (\(p<0.001\)) are found between iFeST and other methods using paired t-tests.}
\label{tab:3}
\centering
\fontsize{8.5}{10.2}\selectfont
\setlength{\tabcolsep}{3pt}
\renewcommand{\arraystretch}{1.14}
\begin{tabular*}{0.98\textwidth}{@{\extracolsep{\fill}}cccccccc@{}}
\toprule
\multirow{2}{*}[-0.45ex]{\shortstack{Inter-Slice\\Correction}} & \multirow{2}{*}[-0.45ex]{\shortstack{Intra-Slice\\Correction}} & \multicolumn{3}{c}{dMRI Volume} & \multicolumn{3}{c}{FOD} \\
\cmidrule(lr){3-5}\cmidrule(lr){6-8}
 &  & SSIM\(\uparrow\) & NRMSE\(\downarrow\) & PSNR (dB)\(\uparrow\) & SH0-SSIM\(\uparrow\) & DWI-SSIM\(\uparrow\) & ACC\(\uparrow\) \\
\midrule
SVRTK & SVRTK & \(0.5823\pm0.1060\) & \(0.1746\pm0.0330\) & \(15.33\pm1.69\) & \(0.7163\pm0.0790\) & \(0.7149\pm0.0792\) & \(0.3156\pm0.0329\) \\
\addlinespace[2pt]
SVR-Net & \multirow{4}{*}{iFeST} & \(0.6619\pm0.0937\) & \(0.1058\pm0.0213\) & \(19.69\pm1.73\) & \(0.8207\pm0.0579\) & \(0.8013\pm0.0654\) & \(0.4789\pm0.0327\) \\
DeepPMT &  & \(0.7054\pm0.0918\) & \(0.0965\pm0.0218\) & \(20.53\pm1.89\) & \(0.8467\pm0.0614\) & \(0.8267\pm0.0684\) & \(0.5367\pm0.0430\) \\
AFFIRM &  & \(0.7086\pm0.0923\) & \(0.0978\pm0.0227\) & \(20.42\pm1.94\) & \(0.8414\pm0.0633\) & \(0.8234\pm0.0695\) & \(0.5383\pm0.0398\) \\
SVoRT &  & \(0.7149\pm0.0912\) & \(0.0951\pm0.0207\) & \(20.65\pm1.84\) & \(0.8501\pm0.0610\) & \(0.8323\pm0.0668\) & \(0.5546\pm0.0413\) \\
\addlinespace[2pt]
\multirow{3}{*}{iFeST} & U-Net & \(0.7316\pm0.0937\) & \(0.0943\pm0.0219\) & \(20.75\pm1.97\) & \(0.8531\pm0.0651\) & \(0.8354\pm0.0696\) & \(0.5602\pm0.0367\) \\
 & PriorNet & \(0.7311\pm0.0946\) & \(0.0945\pm0.0226\) & \(20.74\pm2.03\) & \(0.8548\pm0.0658\) & \(0.8375\pm0.0698\) & \(0.5639\pm0.0384\) \\
 & cGAN & \(0.6842\pm0.0943\) & \(0.1019\pm0.0216\) & \(20.02\pm1.75\) & \(0.8268\pm0.0681\) & \(0.8103\pm0.0711\) & \(0.4971\pm0.0307\) \\
\midrule
iFeST & iFeST & \(\boldsymbol{0.7425\pm0.0956}\) & \(\boldsymbol{0.0934\pm0.0248}\) & \(\boldsymbol{20.88\pm2.21}\) & \(\boldsymbol{0.8614\pm0.0691}\) & \(\boldsymbol{0.8423\pm0.0722}\) & \(\boldsymbol{0.5982\pm0.0492}\) \\
\bottomrule
\end{tabular*}
\end{table*}

The quantitative evaluations of intra-slice motion artifact correction are presented in Table~\ref{tab:2}. The baseline U-Net achieved an SSIM of 0.8630. By incorporating adjacent slices as guidance, PriorNet improved the SSIM to 0.8707, but its overall performance on NRMSE and PSNR deteriorated. In contrast, iFeST achieved the highest SSIM (0.8843), the lowest NRMSE (0.0437), and the highest PSNR (29.89 dB). Fig.~\ref{fig:6} compares corrected images under several types of image corruption. Fig.~\ref{fig:6}(a) illustrates severe signal dropout caused by motion during diffusion sensitization. While competing networks recovered only basic anatomical structures with blurry edges, iFeST produced clearer images with the lowest residual errors. Fig.~\ref{fig:6}(b) shows structural aliasing that competing networks failed to remove. Using frequency-domain modeling, iFeST corrected these k-space anomalies and restored detailed structures. Furthermore, Fig.~\ref{fig:6}(c)--(d) present inputs with low image quality and strong noise. iFeST substantially improved the visual SNR and maintained the lowest global error, confirming its superior performance in correcting diverse intra-slice motion artifacts.

\subsubsection{Volume Reconstruction and FOD Fitting}\label{sec:4.1.3}

\begin{figure*}[!t]
\centering
\includegraphics[width=0.94\textwidth]{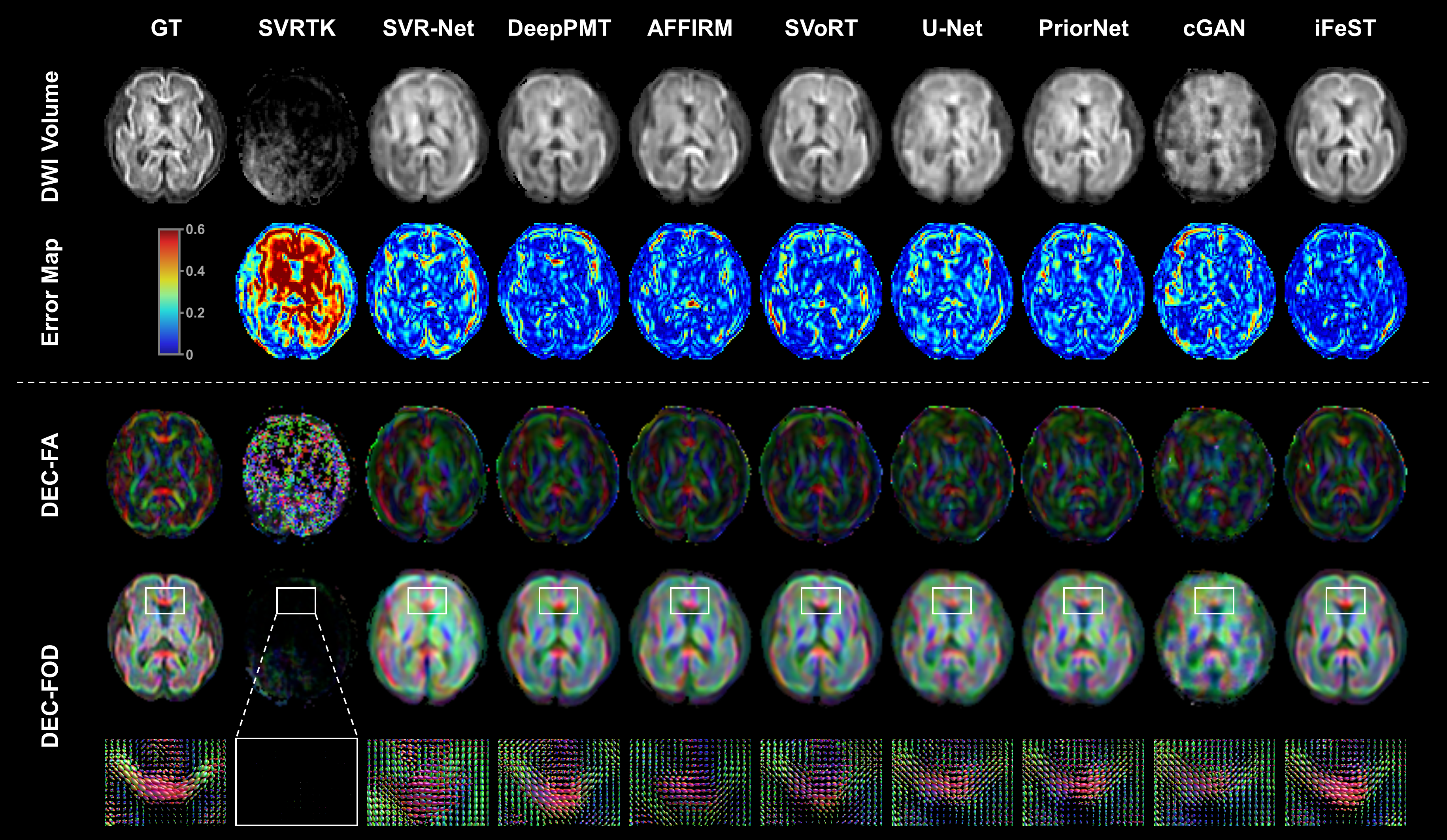}
\caption{Qualitative comparison of dMRI volume reconstruction and FOD fitting on the simulated dataset. The top half shows reconstructed dMRI volumes and error maps. The bottom half presents fitted DEC-FA maps, DEC-FOD maps, and FOD glyphs in the corpus callosum. iFeST best preserved anatomy and white matter orientations, with the lowest residual errors. DEC denotes directionally encoded color.}
\label{fig:7}
\end{figure*}

Table~\ref{tab:3} presents the comparison of reconstruction and FOD fitting quality. iFeST supplied the complementary correction for each single-task network. Constrained by its traditional iterative optimization, the baseline SVRTK struggled with the complex and severe motion, leading to the worst reconstruction (SSIM of only 0.5823). Among the inter-slice MoCo networks, SVR-Net yielded poor reconstruction quality (SSIM = 0.6619), while the other competing networks produced relatively similar results (SSIM ranging from 0.7054 to 0.7149). In the evaluation of intra-slice correction networks, U-Net and PriorNet achieved notably higher reconstruction metrics (SSIM of 0.7316 and 0.7311, respectively). Ultimately, the complete iFeST framework achieved the optimal performance in both volume reconstruction (SSIM = 0.7425) and FOD fitting.

Fig.~\ref{fig:7} presents qualitative comparisons for a 30-week GA fetal brain. Traditional SVRTK exhibited total reconstruction failure, whereas SVR-Net introduced noticeable anatomical deformation errors. Although the three intra-slice networks achieved high quantitative metrics, their reconstructed volumes suffered from over-smoothing, resulting in blurred edges and lower visual SNRs. In contrast, iFeST showed the lowest residual errors and closest structural agreement with GT. In the directionally encoded color FA (DEC-FA) maps, only AFFIRM, SVoRT, and iFeST adequately recovered primary white matter orientations. Magnified FOD glyphs in the corpus callosum (CC) further showed that iFeST most faithfully recovered the expected U-shaped trajectory with red lobes, while competing methods exhibited geometric distortion, lobe attenuation, and orientation deviations.

\begin{table*}[!t]
\caption{dMRI volume reconstruction and FOD fitting on real-world data. Values are reported as mean \(\pm\) standard deviation. Significant differences (\(p<0.001\)) are found between iFeST and SVRTK using paired t-tests.}
\label{tab:4}
\centering
\fontsize{8.5}{10.2}\selectfont
\setlength{\tabcolsep}{3pt}
\renewcommand{\arraystretch}{1.14}
\begin{tabular*}{0.98\textwidth}{@{\extracolsep{\fill}}ccccccc@{}}
\toprule
\multirow{2}{*}[-0.45ex]{Method} & \multicolumn{3}{c}{dMRI Volume} & \multicolumn{3}{c}{FOD} \\
\cmidrule(lr){2-4}\cmidrule(lr){5-7}
 & SSIM\(\uparrow\) & NRMSE\(\downarrow\) & PSNR (dB)\(\uparrow\) & SH0-SSIM\(\uparrow\) & DWI-SSIM\(\uparrow\) & ACC\(\uparrow\) \\
\midrule
SVRTK & \(0.8220\pm0.0950\) & \(0.1073\pm0.0630\) & \(21.70\pm3.27\) & \(0.8278\pm0.0938\) & \(0.8186\pm0.0978\) & \(0.5209\pm0.0740\) \\
iFeST & \(\boldsymbol{0.8495\pm0.0823}\) & \(\boldsymbol{0.0903\pm0.0552}\) & \(\boldsymbol{23.26\pm3.35}\) & \(\boldsymbol{0.8606\pm0.0818}\) & \(\boldsymbol{0.8512\pm0.0857}\) & \(\boldsymbol{0.5555\pm0.0881}\) \\
\bottomrule
\end{tabular*}
\end{table*}

\subsection{Real-World Experiment}\label{sec:4.2}

\begin{figure*}[!t]
\centering
\includegraphics[width=0.90\textwidth]{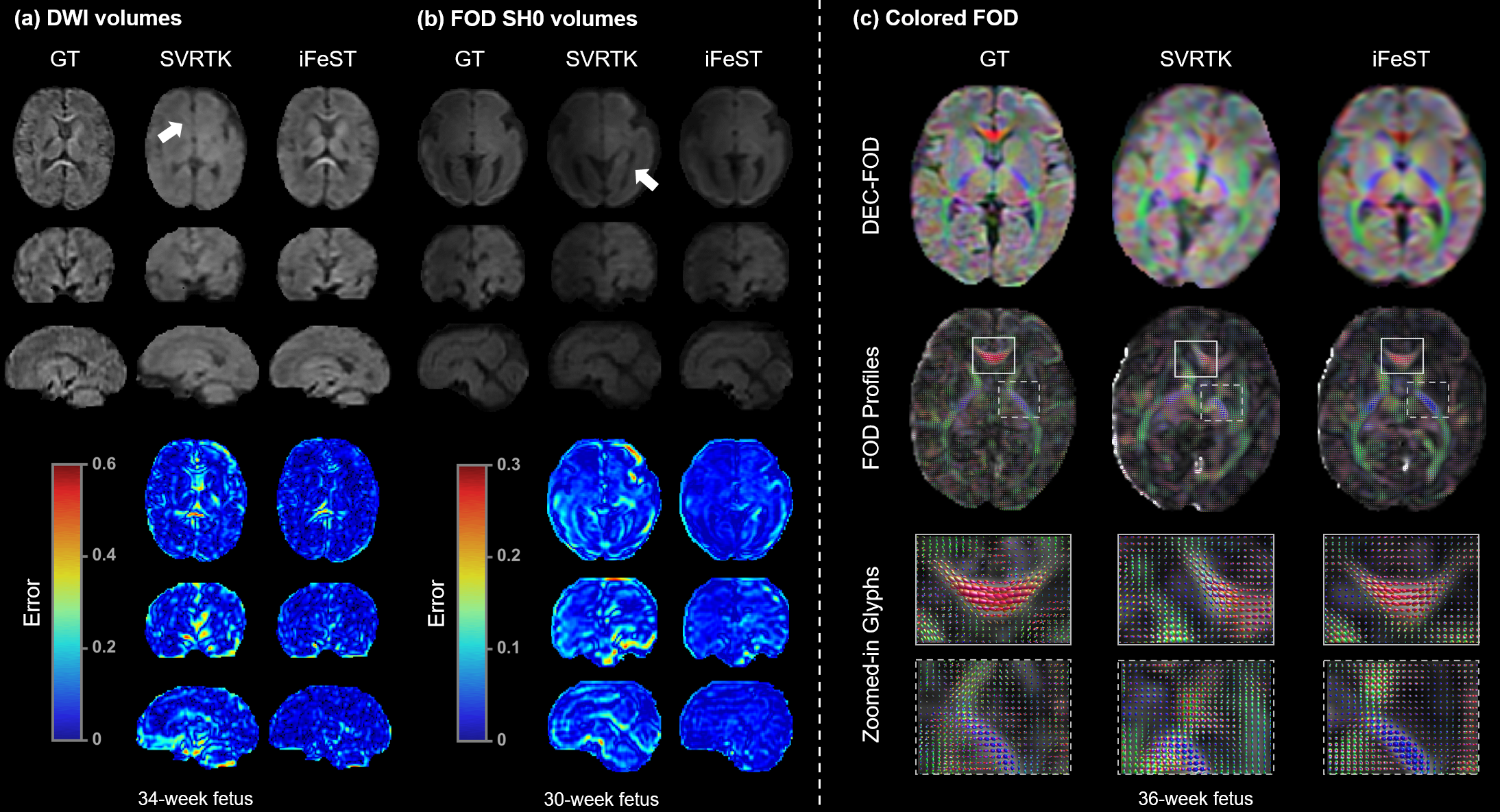}
\caption{Volume reconstruction and FOD estimation on real-world data. (a) and (b) show DW volumes and zero-order FOD coefficient volumes, respectively, with orthogonal views and error maps. White arrows indicate local distortions and blurred boundaries. (c) shows DEC-FOD maps, FOD profiles, and magnified glyphs. White boxes denote regions near the genu and splenium of the corpus callosum. GT denotes the pseudo-GT reference.}
\label{fig:8}
\end{figure*}

To validate the real-world generalizability of the proposed iFeST, we evaluated its performance on 26 real-world fetal brain dMRIs, each containing 4 to 15 motion-corrupted stacks. As shown in Table~\ref{tab:4}, the iFeST-integrated pipeline significantly outperformed the baseline across all metrics (\(p<0.001\), paired t-tests).

\begin{table}[!htbp]
\caption{The ablation results for iFeST and its variants on the simulated dataset.}
\label{tab:5}
\centering
\fontsize{8.5}{10.2}\selectfont
\setlength{\tabcolsep}{3pt}
\renewcommand{\arraystretch}{1.14}
\begin{tabularx}{\columnwidth}{@{}>{\raggedright\arraybackslash}Xcc@{}}
\toprule
\multicolumn{1}{@{}c}{Model Variant} & \shortstack{Motion\\wMAE\(\downarrow\)} & \shortstack{Image\\SSIM\(\uparrow\)} \\
\midrule
full model (iFeST) & \(1.7309\) & \(0.8843\) \\
w/o intra-slice task \& iteration & \(2.2872\) & -- \\
w/o inter-slice task \& iteration & -- & \(0.8710\) \\
w/o iteration & \(2.0301\) & \(0.8776\) \\
w/o frequency modeling & \(1.8090\) & \(0.8786\) \\
w/o slice-wise conv & \(1.9859\) & \(0.8651\) \\
w/o multi-stage conv-embedding & \(1.7702\) & \(0.8811\) \\
w/o subject-specific reference updates & \(2.0279\) & \(0.8771\) \\
w/o motion incremental prediction & \(1.8252\) & \(0.8778\) \\
\bottomrule
\end{tabularx}
\end{table}

Fig.~\ref{fig:8} further provides a qualitative comparison of volume reconstruction and FOD estimation. In reconstructed DW volumes and zero-order FOD coefficient volumes, SVRTK showed residual inter-slice misalignment and local signal distortion, causing blurred boundaries and anatomical distortions (white arrows). In contrast, iFeST reduced these residuals and improved structural consistency with lower errors. The DEC-FOD maps and FOD glyph profiles in Fig.~\ref{fig:8}(c) further showed that iFeST generated more coherent fiber orientations and local FOD profiles closer to the pseudo-GT than SVRTK, particularly in the CC.

\subsection{Ablation Study}\label{sec:4.3}

Lastly, as shown in Table~\ref{tab:5}, we conducted ablation studies on the simulated dataset to validate the effectiveness of the core components within the iFeST framework.

To verify the necessity of multi-task learning and iteration, we first tested non-iterative single-task and dual-task variants. Single-task models yielded a motion wMAE of 2.2872 and image SSIM of 0.8710. Introducing dual-task learning improved wMAE to 2.0301 and SSIM to 0.8776, confirming that inter-slice alignment and intra-slice restoration shared complementary information. Furthermore, the non-iterative dual-task variant underperformed full iFeST, demonstrating that a single-step approach struggled with complex motion.

Second, regarding architecture, removing frequency-domain modeling increased wMAE to 1.8090 and reduced SSIM to 0.8786, indicating that k-space modeling was critical for correcting intra-slice artifacts. Replacing \(3\times 3\times 1\) convolutional kernels with standard 3D convolutions caused over-smoothing along the slice-selection direction, increasing wMAE to 1.9859 and reducing SSIM to 0.8651, the lowest SSIM among all variants. Additionally, feeding only the bottleneck feature to the inter-slice branch without multi-scale embedding yielded wMAE of 1.7702 and SSIM of 0.8811.

Finally, we evaluated iterative reference generation and parameter updates. Replacing the FOD reference with the fixed atlas removed subject-specific guidance, increasing wMAE to 2.0279 and reducing SSIM to 0.8771. Furthermore, removing incremental prediction prevented use of previous motion parameters. Consequently, wMAE rose to 1.8252, and greater structural mismatch between reference and input reduced SSIM to 0.8778.

\section{Discussion}\label{sec:5}

Fetal brain dMRI is susceptible to unpredictable motion that can induce inter-slice misalignment, intra-slice aliasing, and signal dropout. In this work, we proposed iFeST, an iterative framework driven by frequency-spatial-temporal modeling and multi-directional dMRI information integration, to jointly correct intra- and inter-slice motion. Unlike existing approaches that process each directional stack independently and struggle with severe corruption, iFeST incorporates FOD-guided q-space integration within its iterative loop, leveraging directional redundancy from dense dMRI acquisitions to guide correction of corrupted data. Validation on simulated and real-world datasets demonstrated improved MoCo performance and enabled accurate volume reconstruction and microstructural fitting when integrated with existing pipelines.

Conventional retrospective post-processing for fetal brain dMRI comprised two stages: motion correction and SH-based SRR. The initial stage treated DW stacks as independent, using iterative SVR and SRR in a manner similar to sMRI. It first optimized the slice-specific motion parameters, and then reconstructed volumes to serve as the templates for the next registration, repeating this process until convergence. The aligned slices were then mapped into SH space, with contributions of intra-slice motion-corrupted slices adjusted through forward projection. However, optimization-based SVR struggles with large motion. Furthermore, independent correction fails to exploit directional redundancy, while rejecting intra-slice motion-corrupted slices as outliers compromises reconstruction when a large proportion of data is corrupted.

Deep learning has been widely used for MoCo in fetal brain sMRI. SOTA networks typically combine spatial and sequential features for inter-slice estimation, sometimes with iterative 3D structural reference generation. However, comparable fetal dMRI solutions remain scarce, partly because complex motion during DW acquisition is difficult to model and simulate, motivating our dedicated simulator and correction framework. Furthermore, intra-slice aliasing and signal dropout require image restoration in addition to inter-slice alignment, motivating explicit frequency-domain modeling of k-space artifacts.

In this work, we proposed iFeST, the first learning-based retrospective MoCo framework for fetal brain dMRI. The framework combines FeST with microstructure-informed reference generation modules. To leverage data redundancy, we designed a reference-guided iterative strategy within iFeST, which utilized information from other diffusion directions to guide restoration of target motion-corrupted images. FOD fitting integrates multi-directional data and intermediate outputs; forward projection then generates slice-specific references for the next iteration. Within FeST, a dual-stream architecture extracts reference features to explicitly condition correction. Moreover, recognizing that motion during readout introduces k-space inconsistencies, we incorporated frequency-domain operations into spatial-temporal modeling, enabling joint optimization of intra-slice restoration and inter-slice motion estimation. For GT supervision, the FOD-based fetal dMRI simulator generates DW images at arbitrary times through forward projection. It models inter-slice misalignment through spatial transformations and downsampling, intra-slice artifacts through k-space assembly, and diffusion-sensitization motion through signal dropout and additive noise.

We evaluated iFeST against four inter-slice and three intra-slice networks on simulated data. iFeST achieved lower wMAE for inter-slice motion estimation (Table~\ref{tab:1}) and higher SSIM for intra-slice artifact correction (Table~\ref{tab:2}), with clearer anatomical detail in slices affected by signal dropout or aliasing (Fig.~\ref{fig:6}). Furthermore, the ablation studies showed that both iterative correction and FOD-based reference generation contributed to these gains (Table~\ref{tab:5}). These improvements translated into better DW volume reconstruction and FOD fitting (Table~\ref{tab:3} and Fig.~\ref{fig:7}). We then evaluated iFeST on real-world motion-corrupted data from 26 fetuses. With SVRTK alone, residual inter-slice misalignment and local signal distortion blurred anatomical boundaries. After iFeST correction, the reconstructed DW volumes more closely resembled the pseudo-GT (Table~\ref{tab:4}), and the FOD profiles showed better agreement with the reference fiber orientations, particularly in the corpus callosum (Fig.~\ref{fig:8}).

The present work has several limitations. First, the simulator coarsely approximates intra-slice aliasing and signal dropout and omits eddy currents and motion-induced field inhomogeneities. More rigorous models are therefore needed to reproduce clinical acquisition conditions. Second, regarding computational efficiency, although iFeST deploys all modules on GPU with parallel acceleration, the overall training and inference times remain relatively long. For clinical applications with limited computational resources, inference speed could be improved by reducing the number of iterations. Finally, training and evaluation used single-site, single-shell (b=600 s/mm\textsuperscript{2}) data. Future studies should evaluate and extend generalizability using multi-site, multi-shell datasets.

\section{Conclusion}\label{sec:6}

In this work, we proposed iFeST, an iterative MoCo framework for fetal brain dMRI. To exploit the data inter-dependence of dMRI acquisitions, iFeST utilized FOD fitting to aggregate multi-directional information, thereby generating structural and positional references for motion corrections. Furthermore, by incorporating frequency-spatial-temporal modeling, iFeST enabled the joint optimization of intra-slice artifact restoration and inter-slice motion estimation within a single end-to-end network. Validations on both simulated and real-world datasets demonstrated that iFeST successfully corrected compound motion, establishing a workflow from severely corrupted inputs to microstructural modeling. In summary, this work explored the application of deep learning in fetal brain dMRI motion correction, effectively enhancing the reconstruction quality of motion-corrupted data and improving subsequent fetal neurodevelopmental analysis and diagnosis.

\section*{Acknowledgment}

This work was supported by the National Natural Science Foundation of China (grant nos. 32427802 and U24A20754), the Key Development \& Research Project of Zhejiang Province (2026C02A1262(SD2)), the National Key Research and Development Program (2024YFC2707005), the Postdoctoral Fellowship Program of CPSF (grant no. GZC20251895).

\IEEEtriggeratref{48}

\end{document}